%% file: HeadNeRF.tex
\documentclass[10pt,twocolumn,letterpaper]{article}

\usepackage[pagenumbers]{cvpr} 

\usepackage{graphicx}
\usepackage{amsmath}
\usepackage{amssymb}
\usepackage{bm}
\usepackage{booktabs}
\usepackage{float}
\usepackage{cuted}
\usepackage{capt-of}
\usepackage{overpic}
\usepackage{multirow}
\usepackage{lipsum}
\usepackage{rotating}
\usepackage{gensymb}

\usepackage[pagebackref,breaklinks,colorlinks]{hyperref}

\usepackage[capitalize]{cleveref}
\crefname{section}{Sec.}{Secs.}
\Crefname{section}{Section}{Sections}
\Crefname{table}{Table}{Tables}
\crefname{table}{Tab.}{Tabs.}

\def\cvprPaperID{5362} 
\def\confName{CVPR}
\def\confYear{2022}

\begin{document}

\title{HeadNeRF: A Real-time NeRF-based Parametric Head Model}

\author{\large Yang Hong \quad Bo Peng  \quad Haiyao Xiao \quad Ligang Liu \quad Juyong Zhang\thanks{Corresponding author} \vspace{0.5 mm}\\
{\normalsize University of Science and Technology of China}\\
{\tt\footnotesize \{hymath@mail., pb15881461858@mail., xhy1999512@mail., lgliu@, juyong@\}ustc.edu.cn} \\}


\twocolumn[{
\maketitle

\vspace*{-9mm}

\begin{center}
   \begin{overpic}
        [width=\linewidth]{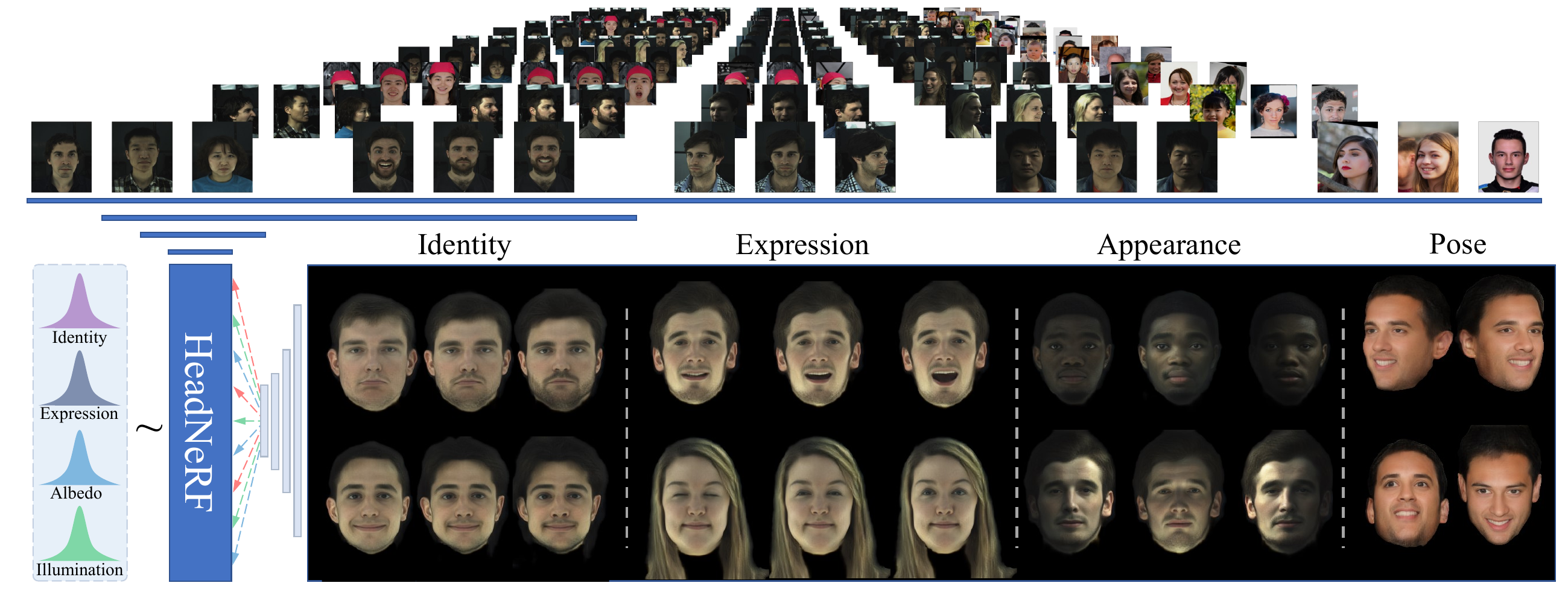}
   \end{overpic}
\end{center}
\vspace*{-8.5mm}
\captionof{figure}{HeadNeRF, a NeRF-based parametric head model, is able to render high fidelity head images in real-time, and supports directly controlling the generated images' rendering pose, and various semantic attributes. The images in the black area are generated by exploring the latent space of HeadNeRF.}
\label{fig:teaser}

\vspace*{5mm}

}]

{
  \renewcommand{\thefootnote}%
    {\fnsymbol{footnote}}
  \footnotetext[1]{Corresponding Author}
}

\begin{abstract}
  In this paper, we propose HeadNeRF, a novel NeRF-based parametric head model that integrates the neural radiance field to the parametric representation of the human head. It can render high fidelity head images in real-time on modern GPUs, and supports directly controlling the generated images' rendering pose and various semantic attributes. Different from existing related parametric models, we use the neural radiance fields as a novel 3D proxy instead of the traditional 3D textured mesh, which makes that HeadNeRF is able to generate high fidelity images. However, the computationally expensive rendering process of the original NeRF hinders the construction of the parametric NeRF model. To address this issue, we adopt the strategy of integrating 2D neural rendering to the rendering process of NeRF and design novel loss terms. As a result, the rendering speed of HeadNeRF can be significantly accelerated, and the rendering time of one frame is reduced from 5s to 25ms. The well designed loss terms also improve the rendering accuracy, and the fine-level details of the human head, such as the gaps between teeth, wrinkles, and beards, can be represented and synthesized by HeadNeRF. Extensive experimental results and several applications demonstrate its effectiveness. The trained parametric model is available at \href{https://github.com/CrisHY1995/headnerf}{https://github.com/CrisHY1995/headnerf}.

\end{abstract}


\section{Introduction}
\label{sec:intro}
\input{"tex_files/Sec01_Introduction.tex"}

\section{Related Work}

\input{"tex_files/Sec02_RelateWork.tex"}

\section{Method}

\input{"tex_files/Sec03_Method.tex"}

\section{Experiments}
\input{"tex_files/Sec04_Experiments.tex"}

\section{Limitation and Future Work}
\label{sec:limitation}
\begin{figure}[t]
  \centering
  \vspace*{1mm}
  \begin{overpic}
      [width=\linewidth]{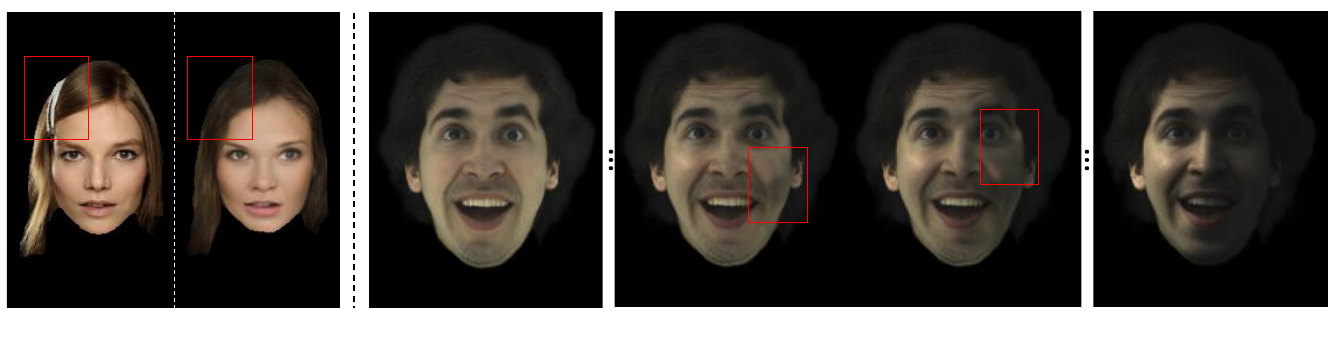}
      \put(0, 0.5){\footnotesize (a) Input}
      \put(15.5, 0.5){\footnotesize Fitting}
      \put(27, 0.5){\footnotesize (b) Continuously change the illumination latent code}
  \end{overpic}
  \vspace*{-5.5mm}
  \caption{Failure results of fitting and re-illumination.}
  \label{fig:limitation}
  \vspace*{-5.5mm}
\end{figure}

There still exist some limitations in HeadNeRF. Although images from FFHQ dataset are added to enhance the representation ability of HeadNeRF, the current training dataset is still not enough to cover various cases. For images that are quite different from our training data, HeadNeRF can only return similar results for the fitting task. As the examples shown in Fig.~\ref{fig:limitation}, because our training data rarely involves images with headgear, it is difficult to render the content of the headgear in the fitting results of HeadNeRF. In the future, we consider using a large amount of in-the-wild face image data to further enhance the representation ability of HeadNeRF in a self-supervised manner.

The current training dataset only contains four types of illuminations in our multi-view dataset~(FaceSEIP), which is insufficient for the coverage of illumination types. Therefore, when we edit the illumination attribute, the change of image shading is not like a continuous movement of the light source position. In the future, we can consider adding data captured by the light stage to alleviate this problem.

\section{Conclusion}
In this paper, we proposed HeadNeRF, a novel NeRF-based parametric head model that integrates neural radiance field to parametric human head representation. Thanks to our well-designed network module and loss terms, HeadNeRF can render high fidelity head images in real-time on modern GPUs and support directly controlling the pose of rendered images and independently editing the identity, expression, and appearance of generated images. Extensive experimental results have demonstrated that HeadNeRF outperforms state-of-the-art related models. We believe that HeadNeRF has taken a significant step toward the realistic digital human.

\noindent{\bf{Ethic discussion.}} Misuse of our methods may raise ethical issues, and the corresponding wrongdoing should be strictly prohibited. Therefore, we require that the media data generated by our method clearly presents itself as synthetic. Moreover, we strictly prohibit using our method to infringe rights, spread misinformation, tarnish reputations, etc. 

\noindent{\bf{Acknowledgements}.} This research was supported by the National Key R\&D Program of China (2020YFC1523102), the National Natural Science Foundation of China (No.62122071, 62025207), the Youth Innovation Promotion Association CAS (No. 2018495) and the Fundamental Research Funds for the Central Universities (No. WK3470000021).

{\small
\bibliographystyle{ieee_fullname}
\bibliography{egbib}
}

\end{document}

%% file: tex_files/Sec01_Introduction.tex
The parametric face/head model, which encodes the human face/head in low-dimensional space, is a hot research topic in computer vision and computer graphics and widely used in many applications like identity recognition~\cite{masi2018deep, wang2021deep}, face analysis~\cite{zollhofer2018state,egger20203d} and film/game production~\cite{song2020accurate}, etc. Early works of the parametric face/head model~\cite{blanz1999morphable, egger20203d, cao2013facewarehouse, paysan20093d,li2017learning, yang2020facescape} mainly model 3D faces with the topologically uniformed face template mesh and usually ignore to represent the non-face parts, such as hair and teeth. With the development of deep learning, 2D generative adversarial networks~(GAN)~\cite{karras2019style, karras2020analyzing} are able to directly render photo-realistic face images without the help of 3D modeling. Some methods~\cite{buhler2021varitex, KowalskiECCV2020, deng2020disentangled, nguyen2019hologan, ghosh2020gif} further introduce semantically disentangled constraints to render the face images in a user-controlled way. However, their rendered results from different views often tend to be inconsistent as they do not explicitly encode or model 3D geometry.

Recently, Mildenhall~\etal~\cite{mildenhall2020nerf} propose to represent 3D scenes using neural radiance fields~(NeRF). This strategy can synthesis photorealistic images and has emerged as a compelling technique. Compared with the above-mentioned generative methods, the density field from NeRF actually implicitly encodes the 3D geometry of the scene. Therefore, the results from NeRF have excellent multi-view consistency. In another opinion, NeRF itself can be regarded as a novel 3D representation equivalent to a textured mesh and naturally supports differentiable rendering. 

Based on this observation, we apply the NeRF structure to the representation of human heads and propose a novel NeRF-based parametric head model, HeadNeRF. HeadNeRF inherits the excellent properties of NeRF, which can generate high fidelity head images and maintain remarkable multi-view consistency. Moreover, NeRF itself supports freely changing the camera perspective used for rendering, and thus HeadNeRF naturally supports the pose editing of rendered images. 
It is challenging for the above-mentioned 2D generative methods. In addition, HeadNeRF intrinsically supports differentiable rendering. In contrast, traditional 3D representations, such as mesh, point cloud, voxel, etc., need to design various approximation methods~\cite{kato2018neural,Laine2020diffrast,jiang2020sdfdiff,liu2019soft,lassner2021pulsar} to alleviate the non-differentiable problems of their rendering process. Therefore, compared with the previous methods, which need to capture and process a large amount of high-quality 3D scan data, the construction of HeadNeRF only needs 2D images as input. Specifically, we collect and process three large-scale human head image datasets and design novel loss terms to disentangle this parametric representation. With the well designed network structure and loss function, HeadNeRF can semantically disentangle the identity, expression, and appearance of the rendered images. Fig.~\ref{fig:teaser} shows some results by freely exploring within the space of HeadNeRF.

We further integrate the volume rendering of NeRF with 2D neural rendering to achieve real-time rendering. Similar with GIRAFFE~\cite{niemeyer2021giraffe} and StyleNeRF~\cite{gu2021stylenerf}, this coarse to fine strategy significantly accelerates the rendering speed of HeadNeRF, and it can exceed 40fps without sacrificing the rendering quality. Benefiting from its nice disentangled representation, real-time rendering of inference, and high fidelity generated results, we apply HeadNeRF to various applications, including novel view synthesis from a single face image, semantically editing face attributes, and even facial reenactment where the expressions of one person are transferred to another person. In summary, the main contributions of this paper include:
\begin{itemize}
    \item We propose the first NeRF-based parametric human head model, which can directly efficiently control the rendering pose, identity, expression, and appearance.
    \item We propose an effective training strategy to train the model from general 2D image datasets, and the trained model can generate high fidelity rendered images.
    \item We design and implement several novel applications with HeadNeRF, and the results verify its effectiveness. 
    We believe that more interesting applications can be explored with our HeadNeRF.
\end{itemize}

\begin{figure*}
    \centering
    \includegraphics[width=\textwidth]{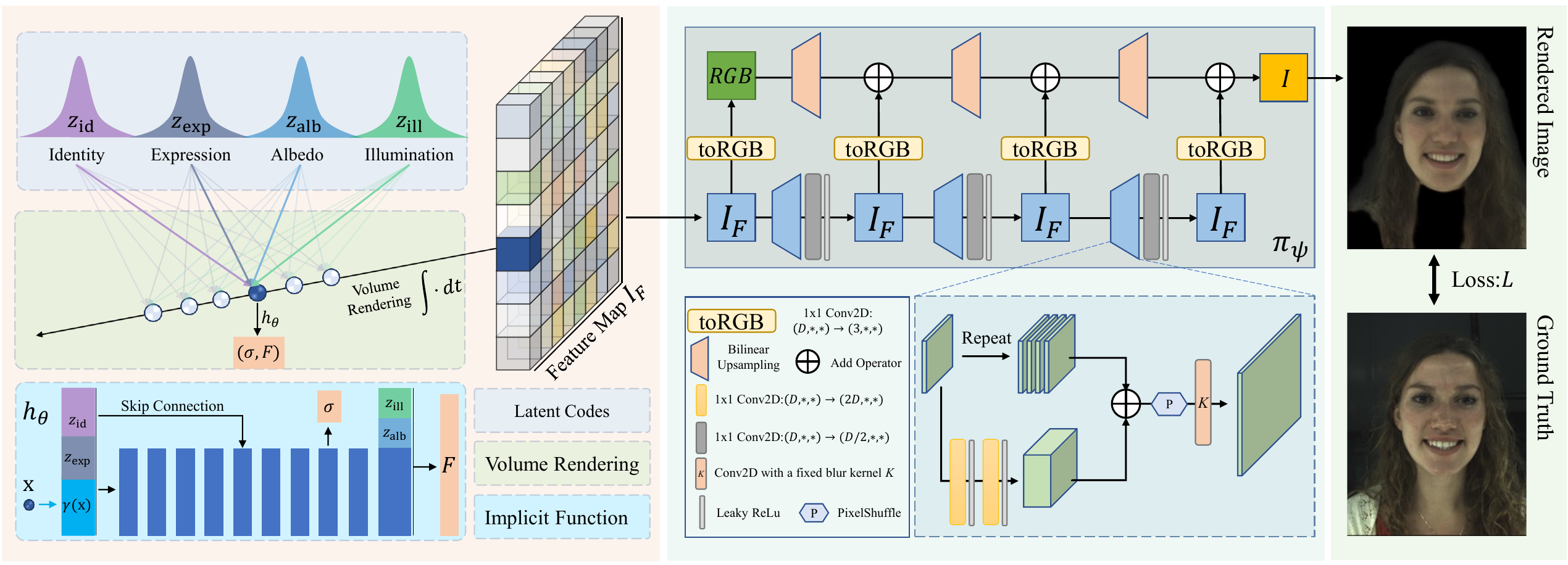}
    \vspace*{-7.5mm}
    \caption{Overview of HeadNeRF. Given semantic latent codes and camera parameters, the MLP-based implicit function $h_\theta$ is utilized to predict the density $\sigma(\mathbf{x})$ and feature vector $F(\mathbf{x})$ of the 3D point $\mathbf{x}$ sampled from one ray. Then we perform volume rendering to generate a low-resolution feature map $I_F$, which is further used to render the final result $I$ by our well-designed 2D neural rendering module $\pi_{\psi}$. The whole process is differentiable, and thus the construction of HeadNeRF can be completed using only 2D images.}
    \label{fig:pipeline}
    \vspace*{-5.mm}
\end{figure*}


%% file: tex_files/Sec02_RelateWork.tex
\noindent{\bf{Parametric Face/Head Model.}} As different people share similar face shapes and appearances, human face can be embedded into a low-dimensional parametric space. Based on this observation, Blanz and Vetter~\cite{blanz1999morphable} propose to build 3D morphable model~(3DMM), which has been further improved and widely used for 3D face representation~\cite{egger20203d, cao2013facewarehouse, paysan20093d,tran2018nonlinear}. 3DMM decomposes the intrinsic attributes of the face into identity, expression, and reflectance. They are encoded as low-dimensional vectors and can be used to restore 3D textured face mesh using corresponding blendshapes. The scene illumination is often simultaneously modeled via the spherical harmonics function~\cite{green2003spherical}. Recently, some generative adversarial~\cite{karras2019style,karras2020analyzing} methods can directly generate photorealistic face images without the aid of 3D modeling. However, these methods pay more attention to generating images that meet the specified distribution and lack semantic and interpretable control over the image synthesis. Some strategies, such as adding disentangled loss~\cite{deng2020disentangled, tewari2020stylerig,KowalskiECCV2020}, embedding the above-mentioned parametric face model into generative adversarial network~(GAN)~\cite{ghosh2020gif,buhler2021varitex}, etc., are proposed to alleviate this problem. However, these 2D GAN methods essentially still lack an explicit 3D geometric structure. Therefore, the rendering results tend to be inconsistent as the camera pose changes.

\noindent{\bf{Neural Radiance Field.}} NeRF~\cite{mildenhall2020nerf} represents 3D scenes using an implicit MLP-based function and shows convincing rendering quality for the novel view synthesis task. Meanwhile, the rendering process of NeRF is inherently differentiable. Therefore, the training of NeRF can be completed only using multi-view images with camera parameters. Benefiting from these advantages, NeRF has been widely used in many fields, such as 3D modeling~\cite{yariv2021volume, wang2021neus}, human face/body digitization~\cite{peng2021neural, su2021nerf, wang2021learning, gafni2021dynamic, guo2021ad, raj2020pva}, generating 4D free-view video~\cite{li2021neural,park2021hypernerf}, etc. Besides, many works are proposed to further improve NeRF, including speeding up training~\cite{barron2021mip, bergman2021fast} and inference~\cite{garbin2021fastnerf, yu2021plenoctrees, wizadwongsa2021nex,hedman2021baking, reiser2021kilonerf}, improving rendering quality~\cite{wang2021learning}, and reducing the number of required inputs~\cite{yu2021pixelnerf,chibane2021stereo}. Please refer to~\cite{tewari2021advances} for a complete summary.

\noindent{\bf{NeRF-Based GANs.}} Some works~\cite{schwarz2020graf,niemeyer2021giraffe,gu2021stylenerf,zhou2021CIPS3D,chan2021pi} integrate NeRF with GANs to design 3D-aware generators. Thanks to the introduction of NeRF structure, these methods generally can directly control the pose of synthesized results and effectively improve the multi-view consistency of generated images, which is challenging for 2D generative models~\cite{karras2019style,karras2020analyzing}. Voxel-based GANs~\cite{nguyen2019hologan, nguyen2020blockgan} can alleviate this problem, but their generated results tend to lack fine details due to the voxel resolution restriction. GRAF~\cite{schwarz2020graf} is the first to introduce the neural radiance field into GAN. Although the quality and 3D consistency of the rendering results can be improved by using this method, it still struggles to render high-resolution and high-fidelity images due to the expensive rendering process of the neural radiance field. GIRAFFE~\cite{niemeyer2021giraffe} further improved the training and rendering efficiency by combining NeRF with a 2D CNN-based neural renderer and can significantly improve the computational speed of NeRF with a slight loss of accuracy. In addition, some works~\cite{chan2021pi,gu2021stylenerf,zhou2021CIPS3D} attempt to design novel ways of applying conditions to generate more fine details.

%% file: tex_files/Sec03_Method.tex
In this work, we propose HeadNeRF, a novel parametric model integrating neural radiance field to human head representation. Unlike the previous 3D mesh-based topologically uniformed face parametric model~\cite{blanz1999morphable,li2017learning,booth2018large,yang2020facescape,cao2013facewarehouse,tran2018nonlinear,paysan20093d}, HeadNeRF takes the accelerated variant of neural radiance field as a unified 3D proxy, which allows it to directly control the viewing pose of the rendered result and generate high fidelity head images in real-time on modern GPUs. To train HeadNeRF, we collected and processed three large-scale face image datasets. Meanwhile, the novel network structure and loss terms are well designed such that the trained parametric model can semantically control and edit the rendering result's identity, expression, and appearance. 

\subsection{Model Representation}
HeadNeRF is a NeRF-based parametric model, denoted as $\mathcal{R}$, which can render an image $I$ with specified attributes for a given camera parameter and some semantic codes. It is formulated as:
\begin{equation}
    I=\mathcal{R}(\mathbf{z}_\textrm{id}, \mathbf{z}_\textrm{exp}, \mathbf{z}_\textrm{alb}, \mathbf{z}_\textrm{ill}, P),
    \label{equ:representation}
\end{equation}
where $P$ is the camera parameter used for rendering, including the extrinsic matrix and the intrinsic matrix. $\mathbf{z}_{*}$ represent the latent codes for four independent factors: identity~$\mathbf{z}_\textrm{id}$, expression~$\mathbf{z}_\textrm{exp}$, the albedo~$\mathbf{z}_\textrm{alb}$ of the face, and the illumination~$\mathbf{z}_\textrm{ill}$ of the scene. 

\subsection{Network Architecture}
Similar to 3DMM, we consider that the underlying geometric shape of the head image is mainly controlled by latent codes related to identity and expression, and the latent codes of albedo and illumination are responsible for the appearance of rendered heads. Therefore, the MLP-based implicit function $h_{\theta}$ of NeRF is adjusted as:
\begin{equation}
        h_{\theta}: 
        (\gamma(\mathbf{x}), \mathbf{z}_\textrm{id}, \mathbf{z}_\textrm{exp},\mathbf{z}_\textrm{alb},\mathbf{z}_\textrm{ill}) \mapsto (\sigma, F),
        \label{equ:facenerf}
\end{equation}
where $\theta$ represents the network parameters, and the network architecture of the implicit function is shown in Fig.~\ref{fig:pipeline}.  $\gamma(*)$ is the pre-defined positional encoding from NeRF~\cite{mildenhall2020nerf}. $\mathbf{x} \in \mathbb{R}^3$ is a 3D point sampled from one ray. Like previous works~\cite{niemeyer2021giraffe,gu2021stylenerf, zhou2021CIPS3D}, instead of directly predicting $\mathbf{x}$'s RGB, we predict a high-dimensional feature vector $F(\mathbf{x}) \in \mathbb{R}^{256}$ for the 3D sampling point $\mathbf{x}$. Specifically, $h_{\theta}$ takes as input the concatenation of $\gamma(\mathbf{x})$, $\mathbf{z}_\textrm{id}$, $\mathbf{z}_\textrm{exp}$ and output the density $\sigma$ of $\mathbf{x}$ and an intermediate feature, the latter and $\mathbf{z}_\textrm{alb}$, $\mathbf{z}_\textrm{ill}$ are used to further predict $F(\mathbf{x})$. Thus, the prediction of the density field is mainly affected by the identity and expression code. The albedo and illumination codes only affect the feature vector prediction, which is consistent with our previous description. Similarly, we remove the viewing direction to avoid capturing undesired inconsistencies caused by dataset bias~\cite{gu2021stylenerf, zhou2021CIPS3D}. 


Then a low-resolution 2D feature map $I_F \in \mathbb{R}^{ 256 \times 32 \times 32 }$ can be obtained by the following volume rendering strategy:
\begin{equation}
    \begin{split}
        I_F(r) & = \int_{0}^{\infty} w(t) \cdot F(r(t)) dt, \\
        \textrm{where}~\quad w(t) & = exp(-\int_{0}^{t} \sigma(r(s))ds) \cdot \sigma(r(t)).
        \label{equ:volume_render}
    \end{split}
\end{equation}
$r(t)$ represents a ray emitted from the camera center. Finally, we map $I_F$ to the final predicted image $I\in \mathbb{R}^{3 \times 256 \times 256}$ with a 2D neural rendering module $\pi_{\psi}$, which is mainly composed of 1x1 Conv2D and leaky ReLU~\cite{Maas13rectifiernonlinearities} activation layer to alleviate possible multi-view inconsistent artifacts~\cite{gu2021stylenerf}. $\psi$ denotes the learnable parameters. Similar with the strategy used in StyleNeRF~\cite{gu2021stylenerf}, the resolution of $I_{F}$ is gradually increased through a series of upsampling layers. The upsampling process can be formulated as:
\begin{equation}
    \begin{split}
        \text{Upsample}(X) &= \text{Conv2D}(Y, K) \\
        Y &= \text{Pixelshuffle}(\text{repeat}(X, 4) + \beta_{\zeta}(X), 2),
        \label{equ:feat_upsample_layer}
    \end{split}
\end{equation}
where $\beta_{\zeta}: \mathbb{R}^{D} \rightarrow \mathbb{R}^{4D} $ is a learnable 2-layer MLP, $\zeta$ denotes the learnable weights, and $K$ is a fixed blur kernel~\cite{zhang2019making}. Like GIRAFFE~\cite{niemeyer2021giraffe}, we map each feature tensor to an RGB image and take the sum of all RGB as the final predicted image. The difference is that we use 1x1 convolution instead of 3x3 convolution to avoid possible multi-view inconsistencies~\cite{gu2021stylenerf}. The network architecture of 2D neural rendering module is shown in Fig.~\ref{fig:pipeline}.

\subsection{Latent Codes and Canonical Coordinate}
\label{sec:latencode}
To efficiently train HeadNeRF, we utilize the 3DMM to initialize the latent codes of each image of our training dataset. Specifically, we set the dimensionality of the latent codes of HeadNeRF to be the same with the dimensionality of the corresponding codes from 3DMM~\cite{tran2018nonlinear} and initialize them by solving inverse rendering optimization~\cite{deng2019accurate,wang2021prior} based on the 3DMM model. Although the initial identity code of 3DMM only describes the geometry of the face area~(without hair, teeth, etc.), it will be adaptively adjusted through the backpropagation gradient of training.

\begin{figure}
    \centering
    \begin{overpic}
        [width=\linewidth]{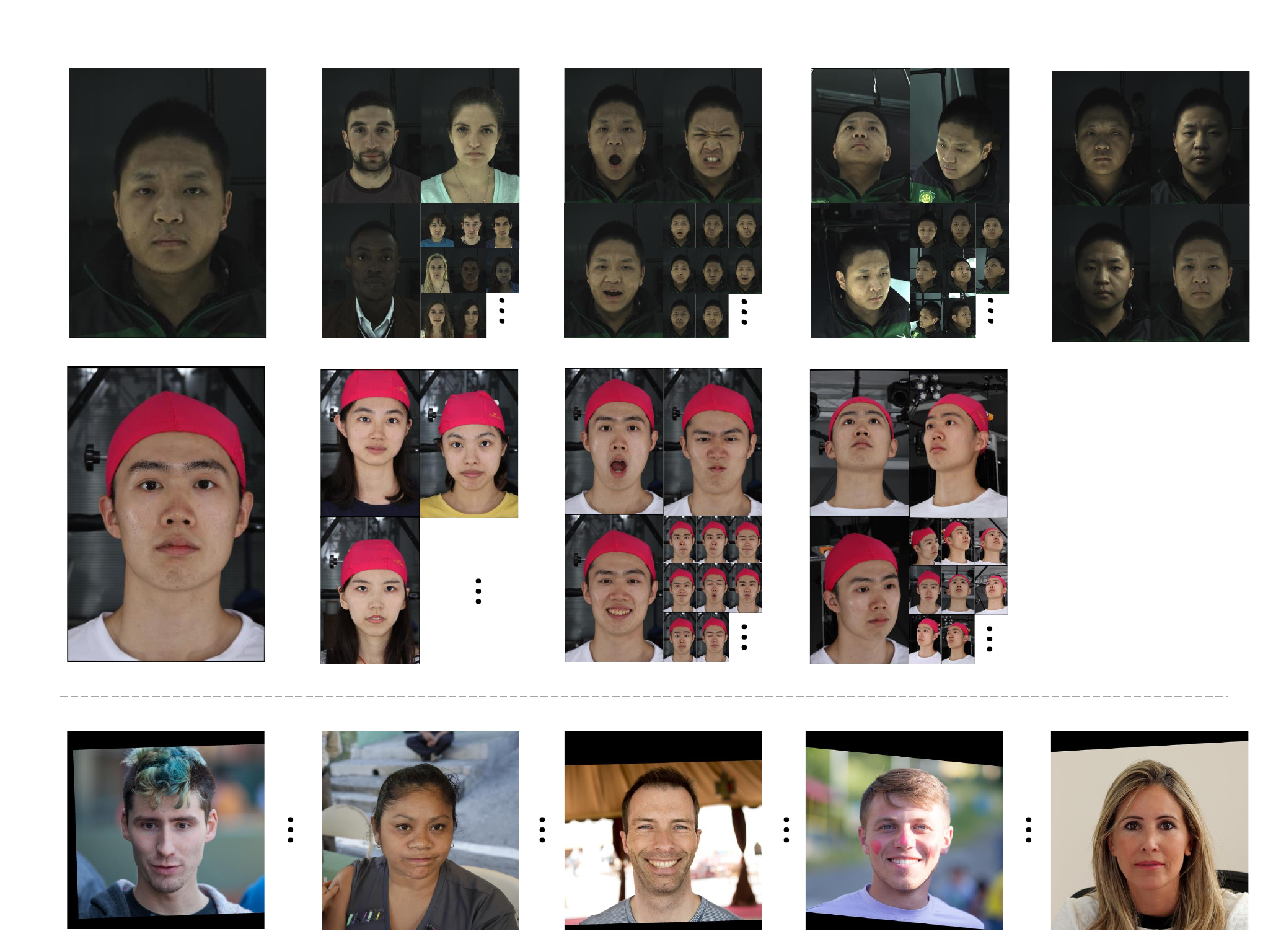}
        \put(28.5, 70){\scriptsize Identity}
        \put(45.3, 70){\scriptsize Expression}
        \put(63.5, 70){\scriptsize Camera pose}
        \put(82.5, 70){\scriptsize Illumination}
        \put(2.5, 63){\begin{turn}{-90} \scriptsize FaceSEIP \end{turn}}
        \put(2, 40.5){\begin{turn}{-90} \scriptsize FaceScape \end{turn}}
        \put(2, 13.5){\begin{turn}{-90} \scriptsize FFHQ \end{turn}}
    \end{overpic}
    \vspace*{-8mm}
    \caption{Some human head images used for training HeadNeRF.}
    \label{fig:dataset}
    \vspace*{-7.0mm}
\end{figure}


On the other hand, the images used for building our model come from different channels~(See sec.~\ref{sec:dataset}). To stabilize the training of HeadNeRF, we need to align each image's underlying geometry to a similar center before training. To this end, for each image, we solve the above-mentioned 3DMM parameter optimization to obtain its corresponding global rigid transformation $T\in \mathbb{R}^{4 \times 4}$, which transforms the 3DMM geometry from 3DMM canonical coordinate to camera coordinate. We further take this transformation as the camera extrinsic parameter of the image. This strategy actually implicitly aligns the underlying geometry of each image to the center of the 3DMM template mesh.

\subsection{Datasets and Preprocessing}
\label{sec:dataset}
We collected and processed three datasets to train HeadNeRF, and the details are described in the following.

\noindent{\bf{FaceSEIP Dataset.}} This dataset includes 51 subjects with different genders, ages, races, and illumination conditions. These subjects are photographed in their daily dress-up and asked to perform 25 specific expressions with 13 cameras and 4 lighting conditions. This dataset contains 66300 face images, and some instances are shown in~Fig.~\ref{fig:dataset}. Besides, the face mask is generated by the off-the-shelf segmentation methods~\cite{MODNet, yu2018bisenet} and the images that fail to segment foreground are manually removed. 

\noindent{\bf{FaceScape Dataset.}} This dataset is from FaceScape~\cite{yang2020facescape} and contains 359 valid subjects. Each subject wears a hood and is asked to perform 20 specific expressions. Since the camera number of each subject is not fixed, for the convenience of training, we select subjects with 10 common perspectives from all subjects and filter out the rest. We also adopt the above-mentioned strategy to generate the face mask for each image. In the final, it contains 124 valid subjects, and some instances are shown in~Fig.~\ref{fig:dataset}.

\noindent{\bf{FFHQ Dataset.}} This dataset is from StyleGAN~\cite{karras2020analyzing} and contains 70000 high-resolution face images. The purpose of using this dataset is to utilize various in-the-wild face images to enhance the generalization ability of HeadNeRF. We also apply the above segmentation strategy to generate the segmentation mask and then manually select 4133 images with good masks. As shown in~Fig.~\ref{fig:ffhq_dataset}, this single-view in-the-wild image dataset effectively improves the generalization ability of HeadNeRF.


\subsection{Loss Function}
All the above-mentioned images are used to train HeadNeRF. The learnable variables include the latent codes of each image and the shared network parameters of volume rendering and neural rendering. The loss terms used to train the model include the following three terms.


\noindent{\bf{Photometric Loss.}} For each image, it is required that the rendered result of the head area to be consistent with the corresponding real image, this loss term is formulated as:
\begin{equation}
    L_{\textrm{data}}  = \| M_{\textrm{h}} \odot (\mathcal{R}(\mathbf{z}_{\textrm{id}}, \mathbf{z}_{\textrm{exp}}, \mathbf{z}_{\textrm{alb}}, \mathbf{z}_{\textrm{ill}},P) - I_{\textrm{GT}}) \|^{2},
    \label{equ:loss_photomatric_loss}
\end{equation}
where $\mathcal{R}(\mathbf{z}_{\textrm{id}}, \mathbf{z}_{\textrm{exp}}, \mathbf{z}_{\textrm{alb}}, \mathbf{z}_{\textrm{ill}}, P)$ is the rendered image, $I_{\textrm{GT}}$ is the corresponding real image, $M_{\textrm{h}}$ is the head mask and $\odot$ indicates Hadamard product operator.

\noindent{\bf{Perceptual Loss.}} Compared with the vanilla NeRF, HeadNeRF can directly predict the color of all pixels in the rendered image via one inference. Therefore, we adopt the perceptual loss~\cite{johnson2016perceptual} in Eq.~\eqref{equ:loss_perceptual_loss} to further improve the image details of the rendered results.
\begin{equation}
    L_{\textrm{per}}  = \sum_i \| \phi_i(\mathcal{R}(\mathbf{z}_{\textrm{id}}, \mathbf{z}_{\textrm{exp}}, \mathbf{z}_{\textrm{alb}}, \mathbf{z}_{\textrm{ill}},P)) - \phi_i(I_{\textrm{GT}})\|^{2},
    \label{equ:loss_perceptual_loss}
\end{equation}
where $\phi_i(*)$ denotes the activation of the $i$-th layer in VGG16~\cite{simonyan2014very} network. As shown in Fig.~\ref{fig:PerceptualLoss}, $L_{\textrm{per}}$ can significantly improve the details of rendered results.

\noindent{\bf{Disentangled Loss.}} In order to achieve semantically disentangled control to the rendered results, we let all images of one subject share the same identity latent code, and the images of the same expression with different lighting conditions and different captured cameras from the same subject share the same expression latent code.


For the purpose of disentangled representation, different subjects with similar expression should have similar expression codes. The initial expression codes generated by the initialization method of~Sec.\ref{sec:latencode} satisfy this requirement thanks to the introduction of 3DMM. Therefore, we require that the learnable expression code cannot be far away from the initial expression code. Other attributes are also constrained similarly. As 3DMM mainly models the face area without hair, teeth, etc., we relax this constraint for these non-expression attributes. Accordingly, the loss term for disentanglement is designed as:
\begin{equation}
    \begin{split}
        L_{\textrm{dis}} = & w_{\textrm{id}} \| \mathbf{z}_{\textrm{id}}  - \mathbf{z}_{\textrm{id}}^0 \|^{2} + w_{\textrm{exp}} \| \mathbf{z}_{\textrm{exp}}  - \mathbf{z}_{\textrm{exp}}^0 \|^{2} + \\
        &w_{\textrm{alb}} \| \mathbf{z}_{\textrm{alb}}  - \mathbf{z}_{\textrm{alb}}^0 \|^{2} + w_{\textrm{ill}} \| \mathbf{z}_{\textrm{ill}}  - \mathbf{z}_{\textrm{ill}}^0 \|^{2},
    \label{equ:loss_disen_loss}
    \end{split}
\end{equation}
where $\mathbf{z}_{*}$ is the learnable latent code and $\mathbf{z}_{*}^0$ is the initial latent code from the 3DMM model. 

In summary, the overall loss of HeadNeRF is defined as: 
\begin{equation}
    L = L_{\textrm{data}} + L_{\textrm{per}} + L_{\textrm{dis}}.
    \label{equ:final_loss}
\end{equation}

%% file: tex_files/Sec04_Experiments.tex
\subsection{Implementation Details}

We implement HeadNeRF with Pytorch~\cite{paszke2019pytorch} and the learnable parameters are updated using Adam solver~\cite{kingma2014adam} on 3 NVIDIA 3090 GPUs. The sizes of different latent codes are $\mathbf{z}_\textrm{id} \in \mathbb{R}^{100}$, $\mathbf{z}_\textrm{exp}\in \mathbb{R}^{79}$, $\mathbf{z}_\textrm{alb}\in \mathbb{R}^{100}$, and $\mathbf{z}_\textrm{ill}\in \mathbb{R}^{27}$ respectively. 
In the volume rendering process, 64 points are sampled for each ray. In addition, we remove the hierarchical volume sampling of
NeRF to further speed up inference. As a result, our HeadNeRF is able to render one head image more than 40fps without any other specific acceleration or optimization. The loss weights in~Eq.~\eqref{equ:loss_disen_loss} are set to $w_\textrm{id} = w_\textrm{alb} = w_\textrm{ill} = 0.001, w_\textrm{exp}=0.1$.

\begin{figure}[t]
    \centering
    \vspace*{-2mm}
    \begin{overpic}
        [width=\linewidth]{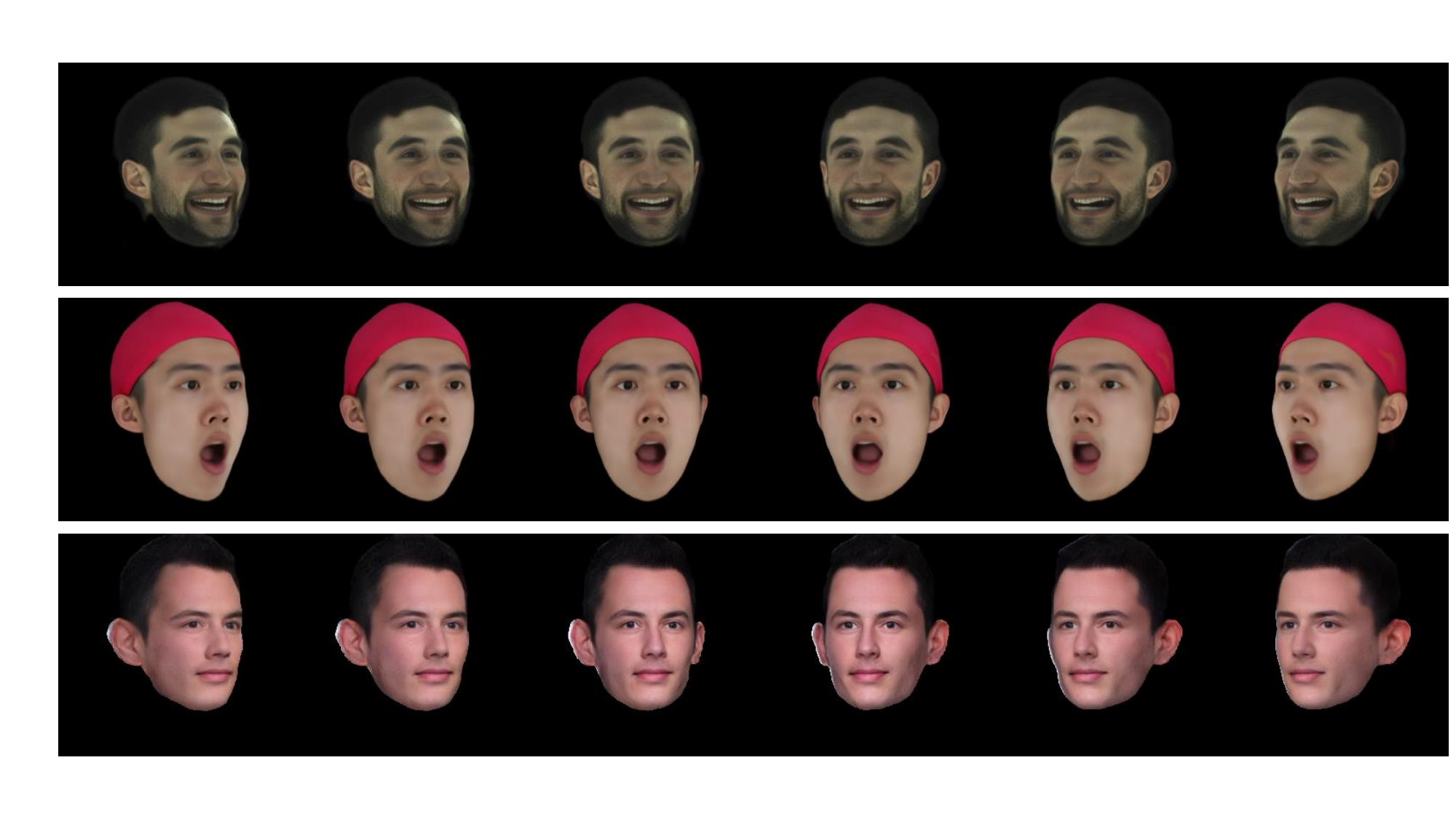}
        \put(0.5,50){\begin{turn}{-90} \scriptsize FaceSEIP \end{turn}}
        \put(0, 33.75){\begin{turn}{-90} \scriptsize FaceScape \end{turn}}
        \put(0, 15.5){\begin{turn}{-90} \scriptsize FFHQ \end{turn}}
    \end{overpic}
    \vspace*{-9.5mm}
    \caption{Disentangled control on camera pose. HeadNeRF can directly control the generated images’ rendering view and synthesize high fidelity rendered results with excellent multi-view consistency.}
    \label{fig:changepose}
    \vspace*{-3mm}
\end{figure}

\begin{figure}[t]
    \centering
    \begin{subfigure}{\linewidth}
      \includegraphics[width=\linewidth]{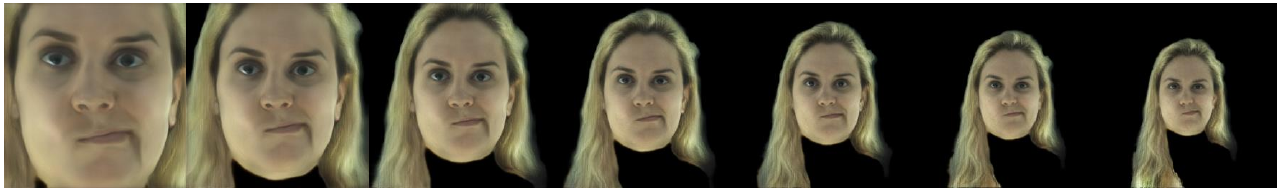}
      \caption{Adjusting camera's position from near to far.}
      \label{fig:adjust_cam_dist}
    \end{subfigure}
    \hfill
    \begin{subfigure}{\linewidth}
        \includegraphics[width=\linewidth]{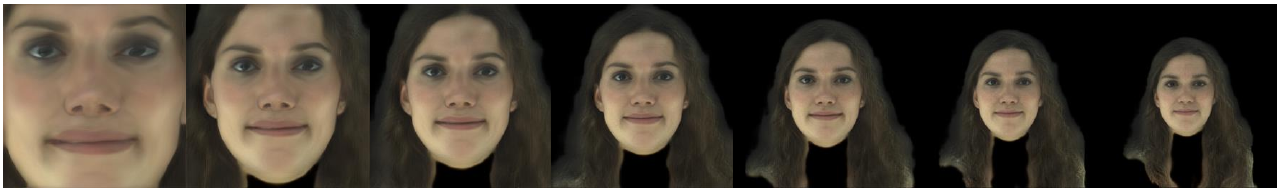}
      \caption{Adjusting camera's FoV from small to large.}
      \label{fig:adjust_cam_fov}
    \end{subfigure}
    \vspace*{-2.5mm}
    \caption{Adjusting camera's position and FoV.}
    \label{fig:adjust_cam_dist_fov}
    \vspace*{-5mm}
\end{figure}

\subsection{Evaluations}
\noindent {\bf{Disentangled Control.}} In this part, we test HeadNeRF's ability to independently control various semantic attributes of rendered results. First, we train HeadNeRF with the datasets mentioned in Sec.~\ref{sec:dataset}. As shown in~Fig.~\ref{fig:changepose},~\ref{fig:adjust_cam_dist_fov}, for a given combination of latent codes $(\mathbf{z}_\textrm{id},\mathbf{z}_\textrm{exp},\mathbf{z}_\textrm{alb},\mathbf{z}_\textrm{ill})$, we can directly adjust the camera parameters to continuously change the rendering view or edit the camera's position and FoV~(Field of View). These rendering results have excellent multi-view consistency and illustrate that our well-designed 2D neural rendering module effectively preserves the geometric structure implicitly encoded by original NeRF.



\begin{figure}[t]
    \centering
    \begin{overpic}
        [width=\linewidth]{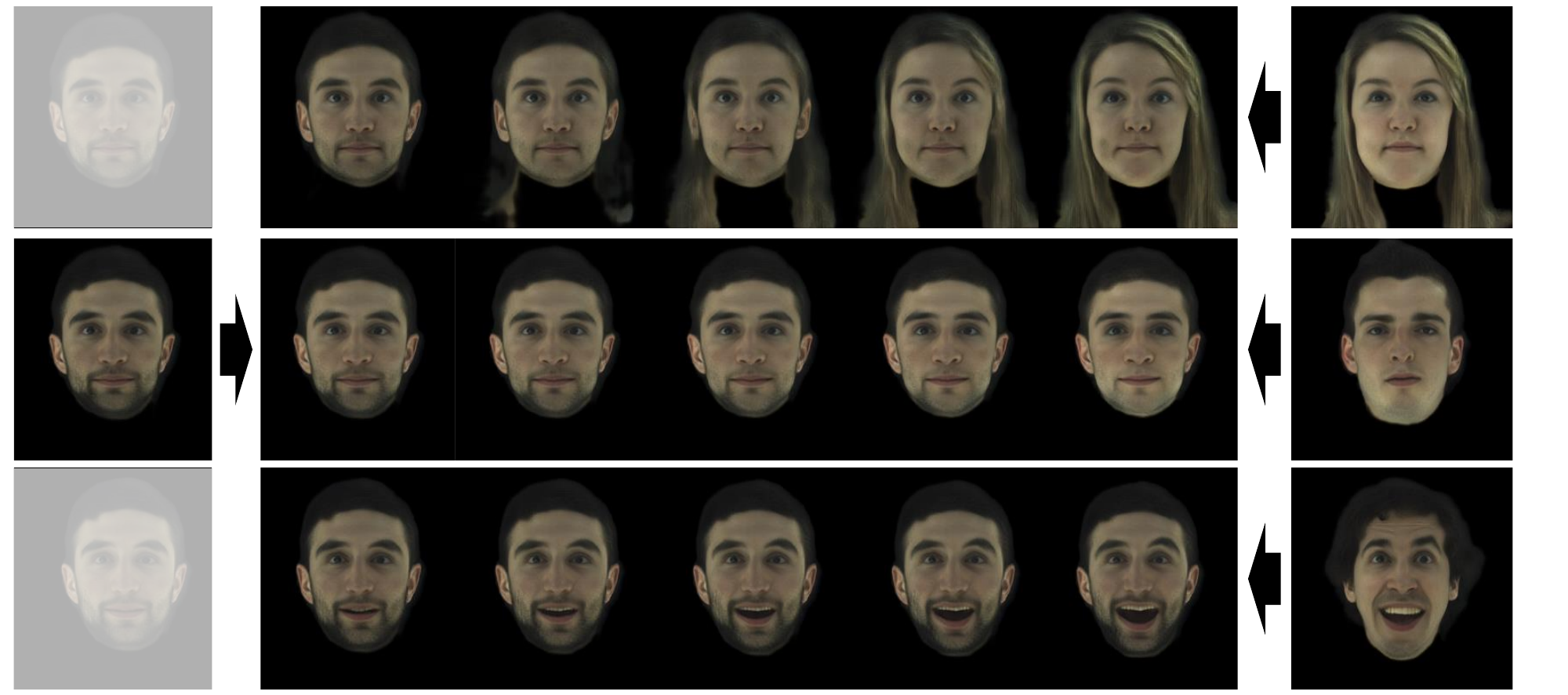}
        \put(97, 41){\begin{turn}{-90} \scriptsize Identity \end{turn}}
        \put(97, 29){\begin{turn}{-90} \scriptsize Appearance \end{turn}}
        \put(97, 13.5){\begin{turn}{-90} \scriptsize Expression \end{turn}}
    \end{overpic}
    \vspace*{-7mm}
    \caption{Disentangle facial attributes. HeadNeRF can independently edit the specific attributes of the rendered result by performing linear interpolating on the latent codes of the specified attributes.}
    \label{fig:edit_code}
    \vspace*{-7.0mm}
\end{figure}

Furthermore, we can utilize HeadNeRF to directly edit semantic attributes of rendered results. As shown in~Fig.~\ref{fig:edit_code}, we first sample several combinations of latent codes and render their corresponding frontal face images. Then two latent codes describing the same attribute are selected and the intermediate latent code can be obtained by performing linear interpolation on them. Finally, we update the interpolation results to the relevant attributes and use HeadNeRF to re-render the head image. As shown in this figure, HeadNeRF can maintain the remaining semantic attributes when editing a specific attribute, which verifies that HeadNeRF effectively disentangles different facial attributes.

\begin{figure}[t]
    \centering
    \vspace*{-0.4mm}
    \begin{overpic}
        [width=\linewidth]{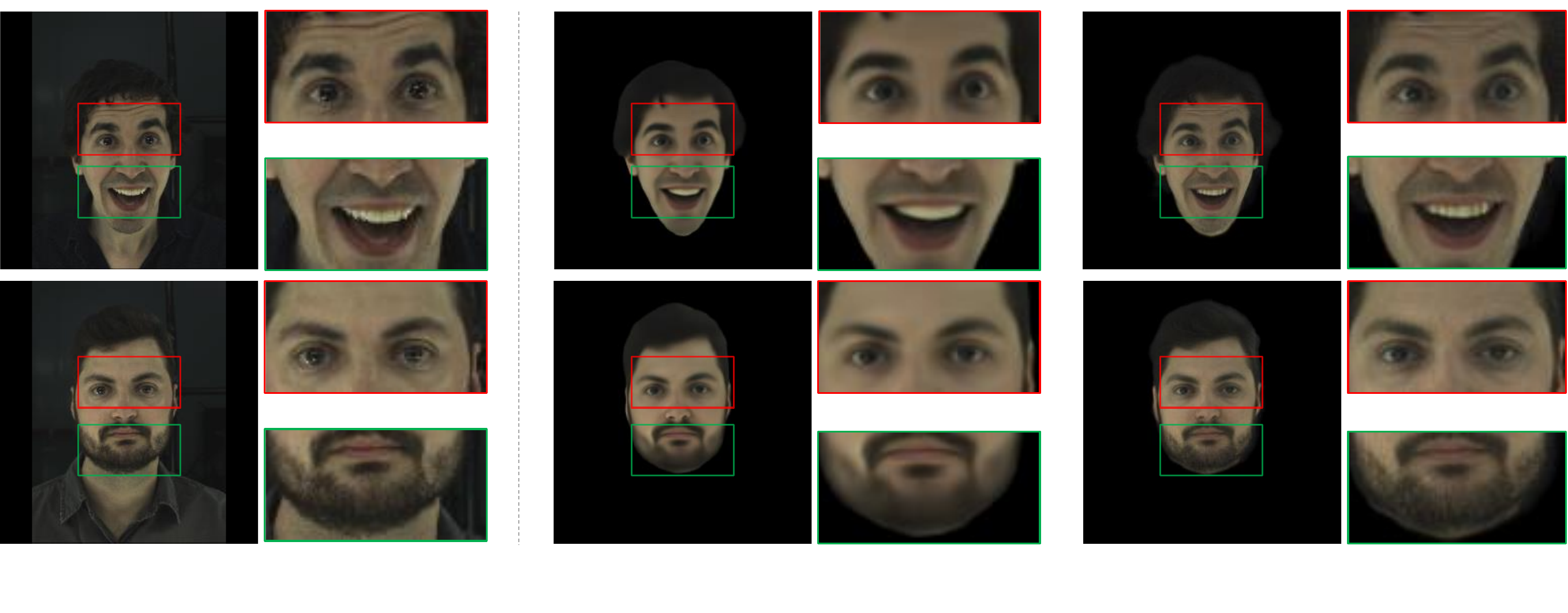}
        \put(5.5, 1.5){ \scriptsize Ground Truth}
        \put(38, 1.5){ \scriptsize w/o Perceptual Loss}
        \put(73, 1.5){ \scriptsize w/ Perceptual Loss}
    \end{overpic}
    \vspace*{-7mm}
    \caption{Ablation study on perceptual loss. The perceptual loss effectively enhances the fine-level details(wrinkle in red area and beard in green area) of the generated results.}
    \vspace*{-2.5mm}
    \label{fig:PerceptualLoss}
\end{figure}

\noindent {\bf{Ablation Study on the  Perceptual Loss.}} We attempt to remove the perceptual loss from the total loss, and record the corresponding trained model as HeadNeRF-noPerc. Then, for a given image in the training dataset, we use the trained HeadNeRF and HeadNeRF-noPerc to generate their prediction results respectively. As shown in~Fig.~\ref{fig:PerceptualLoss}, it can be found that the perceptual loss does effectively enhance the fine-level details of the generated results.

\begin{figure}[t]
    \centering
    \begin{overpic}
      [width=\linewidth]{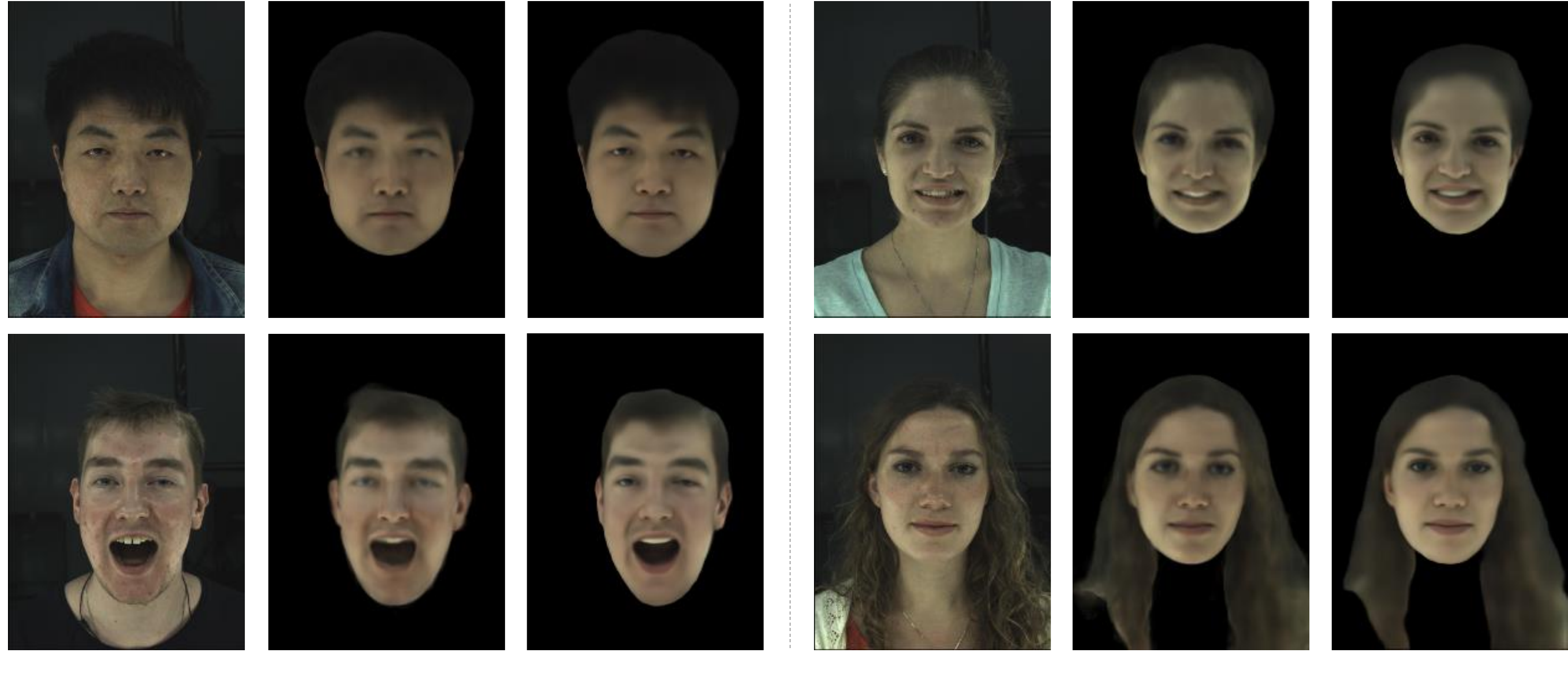}
        \put(0.2, -1.5){\scriptsize Ground Truth}
        \put(18.1, 0){\scriptsize HeadNeRF}
        \put(20.1, -3.25){\scriptsize -vanilla}
        \put(34.5, 0){\scriptsize HeadNeRF}
        \put(36.75, -3.25){\scriptsize w/o $\emph{L}_{\textrm{per}}$}
        \put(51.0,-1.5){\scriptsize Ground Truth}
        \put(69.5, 0){\scriptsize HeadNeRF}
        \put(71.5, -3.25){\scriptsize -vanilla}
        \put(86, 0){\scriptsize  HeadNeRF }
        \put(88.25, -3.25){\scriptsize w/o $\emph{L}_{\textrm{per}}$}
    \end{overpic}
    \vspace*{-3mm}
    \caption{Ablation study on 2D neural rendering. Due to the effectiveness and efficiency endowed by the 2D neural rendering  module, HeadNeRF  can  use  much  less  training time while achieving better rendered results.}
    \label{fig:vsorinerf}
    \vspace*{-5.5mm}
\end{figure}

\noindent {\bf{Ablation Study on 2D Neural Rendering.}} To further verify the effectiveness of 2D neural rendering module in HeadNeRF, we design the following baseline version. All the latent codes, viewing direction and hierarchical sampling are kept except the 2D neural rendering module, and we denote this baseline as HeadNeRF-vanilla. Noting that the baseline can only be trained using the way of sampling batch rays due to the expensive rendering process of NeRF. Thus, the perceptual loss is not available to the HeadNeRF-vanilla. For a fair comparison, our HeadNeRF also removes the perceptual loss. Then, we use FaceSEIP dataset~(See sec.~\ref{sec:dataset}) to train HeadNeRF-vanilla and HeadNeRF, respectively. HeadNeRF-vanilla is trained about 7 days with 3 NVIDIA 3090 GPUs, and HeadNeRF is trained about 3 days with a single NVIDIA 3090 GPU. As shown in~Fig.~\ref{fig:vsorinerf}, the results of HeadNeRF-vanilla tend to be blurred, which may be caused by its inefficient training. In contrast, thanks to the effectiveness and efficiency endowed by the neural rendering module, HeadNeRF can use much less training time while achieving better rendered result. Meanwhile, HeadNeRF-vanilla takes $\sim$5s to render a frame image, while HeadNeRF can render the result in real-time .

\begin{figure*}
    \centering
    \begin{overpic}
      [width=\linewidth]{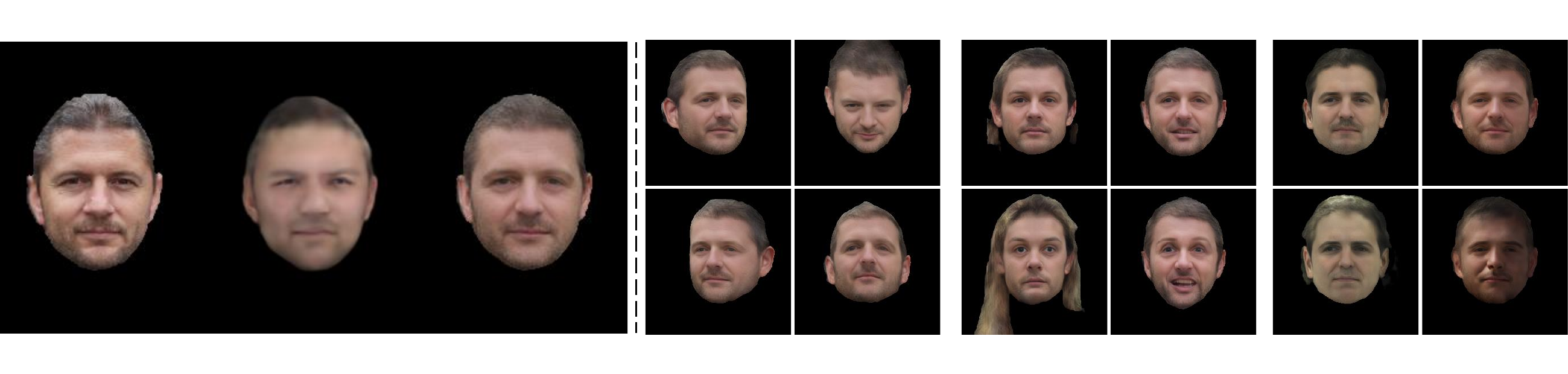}
        \put(4, 0.5){\footnotesize Input}
        \put(16, 0.5){\footnotesize w/o FFHQ}
        \put(30.75, 0.5){\footnotesize w/ FFHQ}
        \put(44.3, 0.5){\footnotesize Yaw}
        \put(53.8, 0.5){\footnotesize Pitch}
        \put(63.5, 0.5){\footnotesize Identity}
        \put(71.8, 0.5){\footnotesize Expression}
        \put(83.5, 0.5){\footnotesize Albedo}
        \put(91.47, 0.5){\footnotesize Illumination}
        \put(17, 22){\normalsize Fitting}
        \put(67.75, 22){\normalsize Editing}
    \end{overpic}
    \vspace*{-7mm}
    \caption{Ablation study on using FFHQ dataset. The introduction of FFHQ dataset significantly improves the generalization ability of HeadNeRF. Based on HeadNeRF, we can modify the specified attributes of the optimization result only with one single image as input, such as adjusting the rendering pose and changing the identity, expression and appearance of the rendered result.}
    \label{fig:ffhq_dataset}
    \vspace*{-6mm}
\end{figure*}

\begin{figure}[thb]
    \centering
    \vspace*{1.1mm}
    \begin{overpic}
        [width=\linewidth]{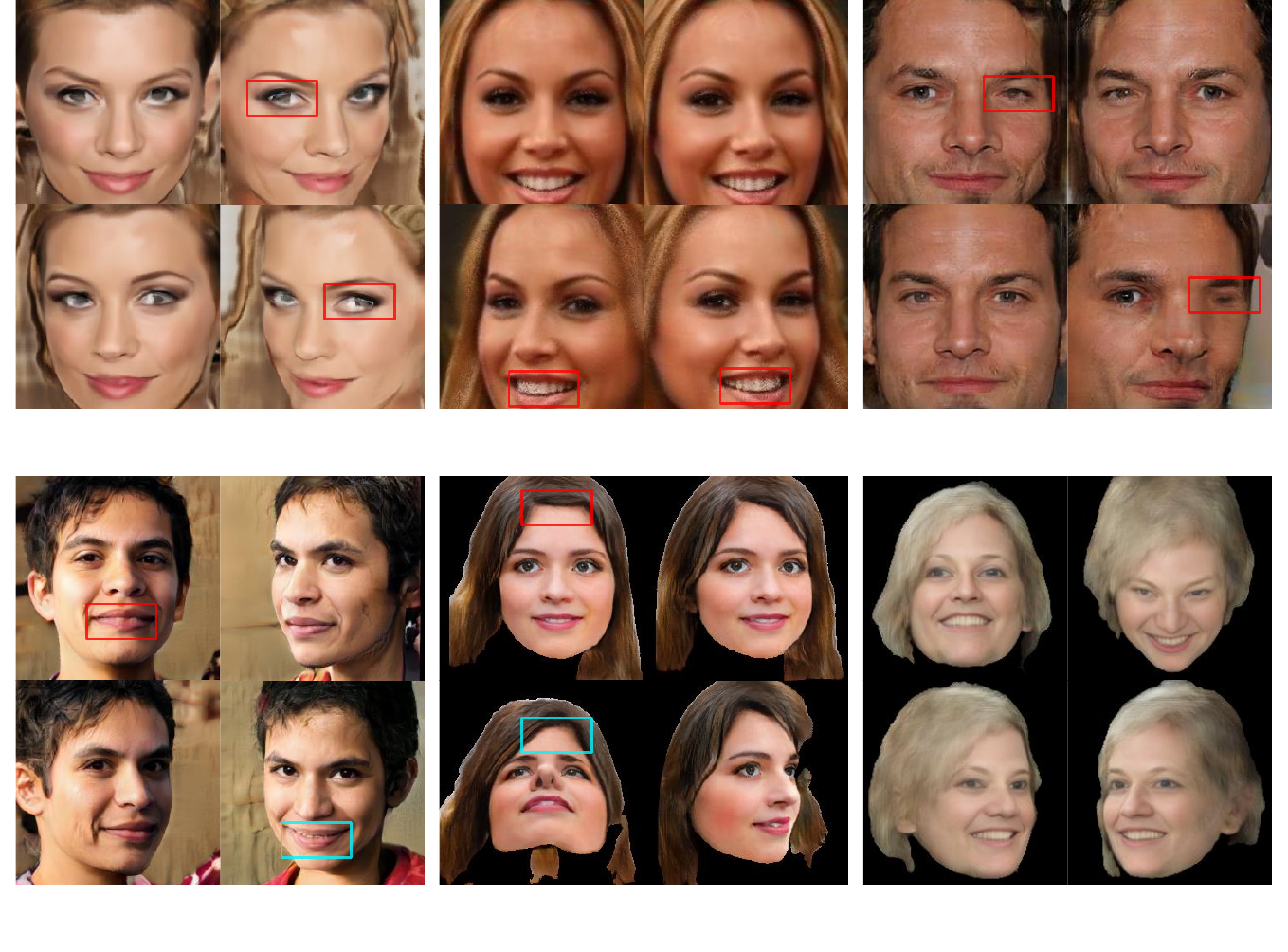}
        \put(10.5, 38.5){\footnotesize GRAF~\cite{schwarz2020graf}}
        \put(43, 38.5){\footnotesize pi-GAN~\cite{chan2021pi}}
        \put(73.5, 38.5){\footnotesize GIRAFFE~\cite{niemeyer2021giraffe}}
        \put(4.5, 1.0){\footnotesize DiscoFaceGAN~\cite{deng2020disentangled}}
        \put(42, 1.0){\footnotesize VariTex~\cite{buhler2021varitex}}
        \put(79, 1.0){\footnotesize Ours}
    \end{overpic}
    \vspace*{-7mm}
    \caption{Qualitative comparison with state-of-the-art methods. The undesired or inconsistent parts of other methods are marked with rectangles. }
    \label{fig:Qualitative_Comparison}
    \vspace*{-5.6mm}
\end{figure}

\noindent {\bf{Ablation Study on Using FFHQ Dataset. }} To verify the importance of FFHQ dataset, we train two parametric models for comparison, where one is only trained with FaceSEIP dataset and the other is trained with FaceSEIP and FFHQ dataset. Next, the network parameters of the two models are fixed, and we further generate the latent codes embedding of the selected image that does not participate in training by minimizing the objective function Eq.~\eqref{equ:final_loss}. The corresponding rendering results of the optimized latent codes are shown in~Fig.~\ref{fig:ffhq_dataset}.
It can be found that the introduction of FFHQ dataset significantly promotes the generalization ability of HeadNeRF. It is worth noting that we can further semantically modify the specified attributes of the optimization result based on our HeadNeRF, such as adjusting the rendering pose and changing the expression, etc. Meanwhile, the driven results are plausible and maintain excellent multi-view consistency. It shows that our HeadNeRF has successfully learned the statistical priors of human head from FaceSEIP dataset so that reasonable semantic editing results can be generated. 
In summary, the datasets used in our model training are complementary and effectively improve the representation ability of HeadNeRF.

\subsection{Comparisons}

\noindent{\bf{Qualitative Comparison.}} Fig.~\ref{fig:Qualitative_Comparison} shows the qualitative comparison between HeadNeRF and some related state-of-the-art methods. Each method generates several face images with different camera poses according to their sampled noise code. Among them, pi-GAN~\cite{chan2021pi}, GRAF~\cite{schwarz2020graf} and GIRAFFE~\cite{niemeyer2021giraffe} are the generative models based on NeRF structure. It can be observed that the generated results of GRAF have obviously inconsistent artifacts, which may be caused by GRAF's inability to globally enforce the rendering results to meet the specified data distribution. Although the images generated by pi-GAN improve visual consistency, they still lack details. Because of the use of discriminant loss and the pure NeRF structure, pi-GAN cannot be trained at high resolution. GIRAFFE also designs a 2D neural rendering module to accelerate the rendering process. However, as shown in this figure, GIRAFFE doesn't get rid of the problem of multi-view inconsistency, which may be caused by the 3x3 CNN in GIRAFFE damaging the 3D information encoded by the NeRF.

It needs to be pointed out that these methods are designed for general objects, and they cannot be used to semantically edit and control various attributes of the generated results. Some other methods like VariTex~\cite{buhler2021varitex} and DiscoFaceGAN~\cite{deng2020disentangled} attempt to integrate the priors from 3DMM into 2D generative models, but their rendering results still suffer from multi-view inconsistencies due to the limited representation ability of 3DMM. In contrast, HeadNeRF can efficiently generate high fidelity images while maintaining excellent multi-view consistency.

\begin{table}[t]
    \begin{center}
    \resizebox{\linewidth}{!}{
    \begin{tabular}{|l|c|c|c|c|c|c|c|c|c|c|}
    \hline
                  & \multicolumn{3}{c|}{CelebAMask-HQ} & \multicolumn{3}{c|}{FFHQ} \\
    \hline
    Method        & L1 $\downarrow$  & PSNR $\uparrow$ & SSIM $\uparrow$ & L1 $\downarrow$  & PSNR $\uparrow$ & SSIM $\uparrow$  \\
    \hline\hline
    pi-GAN~\cite{chan2021pi}&0.543&\textbf{24.8}&\textbf{0.799}&0.483&\textbf{24.9}&	\textbf{0.810}\\
    GIRAFFE~\cite{niemeyer2021giraffe}       &0.420&20.6&0.628&0.428&20.3&	0.635\\
    Ours          &\textbf{0.306}&19.6&0.702&\textbf{0.344}&21.6&0.755\\
    \hline
    \end{tabular}}
    \end{center}
    \vspace{-5mm}
    \caption{Quantitative  comparison  with state-of-the-art generative adversarial models based on NeRF structure. Some metrics about fitting a single image are calculated.}
    \label{tab:Quantitativeevaluation}
    \vspace{-3mm}
\end{table}

\begin{figure}[t]
    \centering
    \vspace*{-0.073mm}
    \begin{overpic}
        [width=\linewidth]{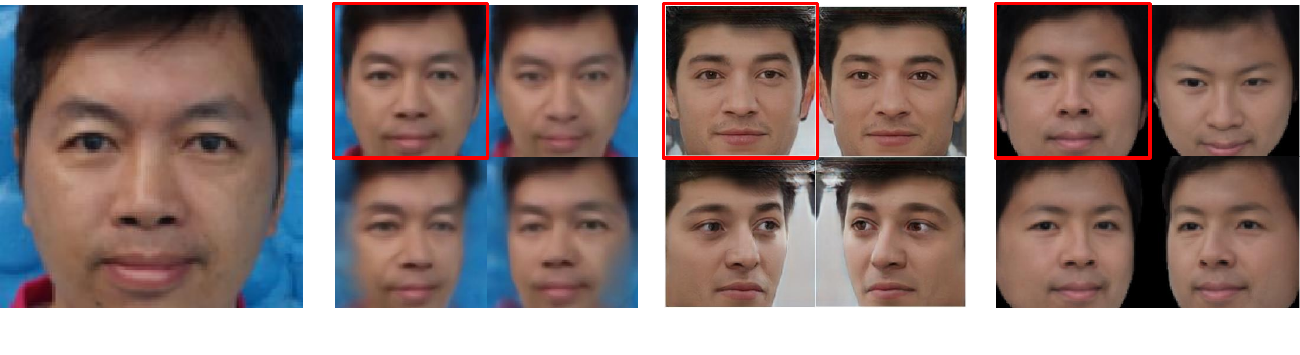}
        \put(7.5, 0){\footnotesize Input}
        \put(30, 0){\footnotesize pi-GAN~\cite{chan2021pi}}
        \put(52.5, 0){\footnotesize GIRAFFE~\cite{niemeyer2021giraffe}}
        \put(86, 0){\footnotesize Ours}
    \end{overpic}
    \vspace*{-6mm}
    \caption{Fitting results of different methods. The figure with red box is the fitting result, and others are generated by changing the rendering pose of the fitting result.}
    \label{fig:Novel_View_Qualitative_Comparison}
    \vspace*{-5mm}
\end{figure}

\begin{figure*}
    \centering
    \begin{overpic}
        [width=\linewidth]{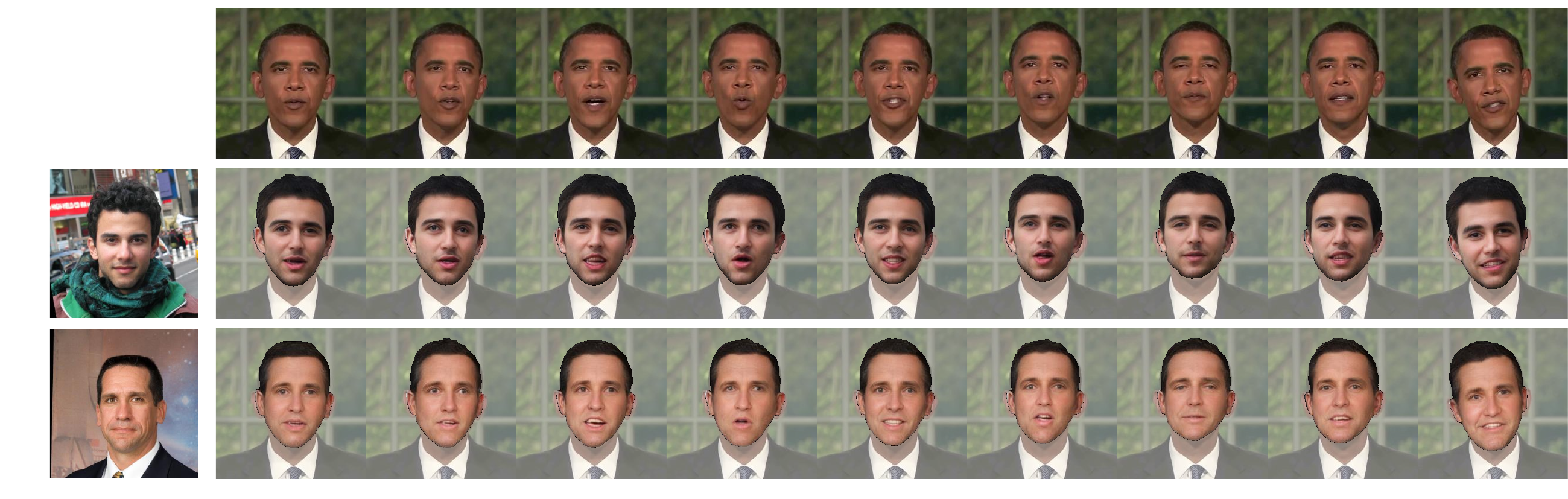}
        \put(3, 26.7){\footnotesize The Reference}
        \put(6, 24.7){\footnotesize Video}
        \put(1, 6){\begin{turn}{90} \footnotesize The Target Image \end{turn}}
    \end{overpic}
    \vspace*{-8.5mm}
    \caption{Expression Transfer. HeadNeRF is used to transfer facial expressions from the reference video to the persons in target images.}
    \label{fig:expression_transfer}
    \vspace*{-5.5mm}
\end{figure*}

\noindent {\bf{Quantitative Comparison.}} Currently, the PSNR~(Peak Signal-to-Noise Ratio) of HeadNeRF on FaceSEIP, FaceScape, and FFHQ datasets are 29.8, 30.6, and 23.3, respectively.  To further quantitatively evaluate the effectiveness of HeadNeRF, we randomly select 400 in-the-wild face images from the CeleAMask-HQ dataset~\cite{CelebAMask-HQ} and the test dataset of FFHQ~\cite{karras2019style}, respectively. These images are used as the evaluation dataset and did not participate in the training of HeadNeRF at all. For each image in the evaluation dataset, the fitting results of pi-GAN~\cite{chan2021pi}, GIRAFFE~\cite{niemeyer2021giraffe} and our HeadNeRF are obtained by solving the inverse rendering optimization. For pi-GAN, the official fitting code and pre-training model are used. For GIRAFFE, we use the official pre-trained model, and the fitting code is implemented by ourselves. Then, the mean $L_1$ distance, the PSNR, and the SSIM~(Structure Similarity Index) is calculated between the input image and fitting results. The statistical results are shown in Tab.~\ref{tab:Quantitativeevaluation}.  

Although the PSNR and SSIM of pi-GAN are optimal, we find that the fitting results of pi-GAN are often prone to overfitting. As shown in Fig.~\ref{fig:Novel_View_Qualitative_Comparison}, the fitting result of pi-GAN is indeed visually optimal. However, if we edit and change the rendering pose of the fitting result, the rendered results often become blurred and damaged. On the other hand, GIRAFFE can generate plausible rendered results when we change the rendering pose of GIRAFFE's fitting result, but the multi-view consistency of the rendered results is undesired. Compared with GIRAFFE, HeadNeRF has higher fitting accuracy and can maintain better multi-view consistency. Please note that our HeadNeRF can further modify and edit other semantic attributes, including identity, expression and appearance. As shown in~Fig.~\ref{fig:ffhq_dataset}. 


\subsection{Application: Expression Transfer}
As HeadNeRF has strong representation ability and can disentangle various attributes of the rendered result, it can be used for many applications such as novel-view synthesis, style mixing, etc,. 
In this part, we utilize HeadNeRF to perform expression transfer, i,e., transfer the facial expressions from the reference video to the people in target picture. To this end, we only need to extract the latent codes of all images from the reference video and the target image, and replace the expression latent code of the target image with the expression latent codes from the reference video. Finally, the trained HeadNeRF is employed to generate the desired face image sequence where the characters in the target image are driven to make expressions from the reference video. The qualitative results are shown in Fig.~\ref{fig:expression_transfer}.

